\documentclass[11pt,a4paper]{article}

\usepackage[T1]{fontenc}
\usepackage[utf8]{inputenc}
\usepackage[english]{babel}
\usepackage{lmodern}
\usepackage{microtype}
\usepackage[a4paper,margin=2.8cm]{geometry}
\usepackage{setspace}
\usepackage{enumitem}
\usepackage[numbers,sort&compress]{natbib}
\usepackage[hidelinks]{hyperref}
\usepackage{url}
\usepackage{fancyhdr}
\usepackage{graphicx}
\usepackage{float}
\title{Conversing with a Conversational Agent:\\ A Real Conversation, Not a Simulated One\\[0.8em]
\large From Objective Simulation to the Actuality of Interaction:\\
Otherness, Individuation, and the Hybrid Semiotic Fabrication of Meaning}
\author{Christophe Denis\\[0.4em]
\normalsize Sorbonne Université\\
\small UMMISCO (UMI 209) --- International Unit of Mathematical and\\
\small Computer Modeling of Complex Systems, Paris, France\\[0.3em]
\normalsize Université Paris 1 Panthéon-Sorbonne\\
\small IHPST (UMR 8590) --- Institute for the History and Philosophy\\
\small of Science and Technology, Paris, France\\[0.4em]
\normalsize \href{mailto:christophe.denis@sorbonne-universite.fr}{christophe.denis@sorbonne-universite.fr}}
\date{}

\begin{document}
\maketitle

% Small illustration immediately after the title, with complete English artwork credit.
\begin{center}
  \vspace{-0.5em}
  \includegraphics[height=0.19\textheight,keepaspectratio]{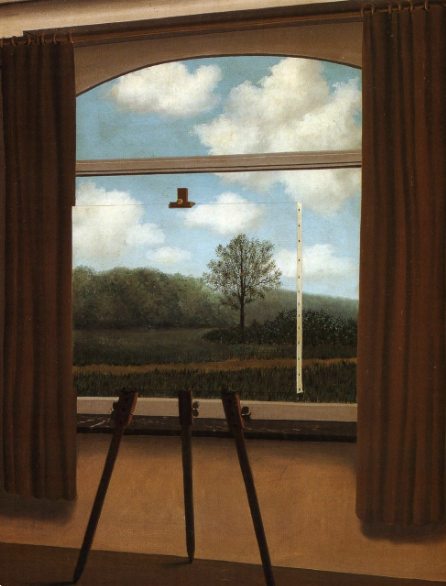}\\[0.35em]
  \begin{minipage}{0.88\textwidth}
  \centering\footnotesize
  René Magritte (1898--1967), \emph{La condition humaine} (\emph{The Human Condition}), 1933. Oil on canvas, 100 $\times$ 81 cm. National Gallery of Art, Washington, D.C. Gift of the Collectors Committee, accession no.\ 1987.55.1.
  \href{https://www.nga.gov/artworks/70170-la-condition-humaine}{Museum catalogue}.
  \end{minipage}
\end{center}
\vspace{-0.6em}

\begin{abstract}
The claim that interactions with artificial conversational agents merely simulate human conversation rests on a philosophical distinction between reproducing the observable features of an activity and actually performing it. Although this distinction is appropriate for certain physical phenomena, its application to linguistic and symbolic activities is far less straightforward. This article argues that conversations with large language model (LLM)-based agents are genuine conversational activities rather than mere simulations, without thereby attributing consciousness, subjective understanding, or human intentionality to artificial systems. Drawing on the distinction between subjective and objective simulation, the article first examines the limitations of defining conversation exclusively through the intrinsic properties of its participants. It then develops a relational and semiotic account of conversational activity, informed by linguistic pragmatics, Lacan's theory of the mirror, Baudrillard's concept of the simulacrum, Simondon's philosophy of individuation, and Peirce's semiotics. These perspectives make it possible to distinguish the reality of conversational interaction from the appearance of artificial subjectivity and from the epistemic validity of the meanings produced. The argument culminates in the concept of Hybrid Semiotic Fabrication, which characterizes meaning-making as a distributed process involving human participants, artificial intelligence, corpora, and technical arrangements. Recognizing the reality of human--AI conversation therefore entails neither anthropomorphizing artificial agents nor granting their outputs epistemic authority. Rather, it requires a more precise account of the relational, semiotic, and normative conditions under which conversational meaning is produced and justified.
\end{abstract}


\section*{Introduction}

The development of large language models (LLMs) has profoundly transformed the ways in which human beings interact with computational systems. Whereas conventional interfaces primarily relied on predefined commands and operations, contemporary conversational agents enable extended exchanges in which users can discuss hypotheses, formulate objections, revise arguments, and progressively develop lines of reasoning. These interactions are no longer confined to the execution of isolated instructions. They increasingly take the form of dialogues whose content and direction evolve through successive contributions from the human participant and the artificial system.

This transformation revives a longstanding philosophical question, already central to Alan Turing's reflections on machine intelligence \cite{turing1950}: does reproducing the characteristic behaviour of a human activity suffice to establish that the activity is actually being performed?

In the case of conversational agents, the answer remains contested. According to one influential position, such systems merely simulate human conversation because they lack consciousness, subjective understanding, and communicative intentionality comparable to those of human beings. The fluency of their responses, their sensitivity to context, and their ability to sustain extended exchanges would therefore amount to outward manifestations of an activity whose essential conditions remain absent.

This objection deserves serious consideration. Yet it also presupposes a particular conception of conversation, one that makes the reality of the interaction dependent on the intrinsic properties of its participants. Conversation can alternatively be approached as a relational activity whose actuality consists in operations of address, response, uptake, contestation, and the transformation of meaning. The relevant question is therefore not merely whether a machine understands what it says, but whether the absence of subjective understanding necessarily disqualifies the conversational activity in which it participates.

The motivation for the present inquiry is to examine this assumption critically. The description of human--AI exchanges as a ``simulation of conversation'' is frequently treated as self-evident, although it presupposes answers to two distinct philosophical questions: what does it mean to simulate an activity, and what constitutes a conversation?

The distinction between subjective and objective simulation, notably examined by St\'ephane Chauvier \cite{chauvier2008}, provides a framework for addressing the first question. The second requires us to consider whether conversation must necessarily be understood as the encounter of two human-like subjectivities or whether it can also be characterized through the relational and semiotic operations through which discourse is constituted.

The central thesis of this article is that conversing with an artificial conversational agent is a genuine conversational activity, not merely the simulation of human conversation.

This thesis does not require us to attribute consciousness to LLMs, to assume that they possess subjective intentionality, or to regard their linguistic productions as necessarily meaningful or true in the same sense as justified human assertions. It requires us to distinguish three questions that are too often conflated: the reality of a discursive interaction, the subjective properties of its participants, and the epistemic status of the meanings produced through that interaction.

The inquiry also has broader implications for the philosophy of technology. If the absence of artificial subjectivity is treated as sufficient evidence for the absence of genuine conversation, we risk overlooking the forms of mediation through which computational systems participate in the organization of contemporary intellectual activities. Conversely, recognizing the reality of these interactions must not lead to an anthropomorphic understanding of machines or to the neglect of the illusions of comprehension and recognition that they may generate. The task is to account for an effective relation without presupposing ontological symmetry between its participants.

The argument develops in two complementary stages.

The first section, ``Conversation and the Paradigm of Simulation,'' examines the philosophical assumptions underlying the claim that conversational agents merely reproduce the appearances of human dialogue. The distinction between subjective and objective simulation clarifies the scope of this argument and reveals the difficulties involved in extending it to symbolic activities.

The second section, ``Conversation as a Relational and Semiotic Reality,'' offers a positive characterization of conversational activity. Pragmatic approaches to interaction help identify its constitutive operations, while Lacan's account of the mirror and Baudrillard's analysis of the simulacrum illuminate the distinction between functional otherness and the appearance of artificial subjectivity. Simondon's philosophy of individuation and Peirce's semiotics then provide resources for understanding conversation as a situated activity constituted through technical mediation and processes of signification.

This analysis ultimately leads to the concept of Hybrid Semiotic Fabrication, through which the reality of human--AI conversation can be acknowledged without presupposing machine consciousness, while preserving the requirements of responsibility, verification, and justification that determine the epistemic standing of the meanings generated in these exchanges.

\section{Conversation and the Paradigm of Simulation}

Whether an exchange with a conversational agent constitutes a genuine conversation raises a philosophical problem that extends well beyond the linguistic performance of artificial systems. It requires us to determine what makes a conversational activity real and under what conditions a machine might participate in it.

According to a widespread conception, conversation necessarily involves properties associated with human subjectivity, including intentionality, understanding, and the capacity to recognize another participant as an interlocutor. Artificial systems might reproduce the observable manifestations of these properties without actually possessing the capacities that confer authenticity upon the corresponding activity. The debate therefore turns on a distinction between simulation and actual performance.

Yet the concept of simulation itself requires clarification. It encompasses philosophically distinct operations, and conflating them risks determining the outcome of the argument before the question has been properly examined.

In ``Simuler et faire simuler,'' St\'ephane Chauvier (2008) distinguishes between subjective simulation and objective simulation \cite{chauvier2008}.

Subjective simulation refers to an act of pretence through which an individual intentionally adopts the appearance of a behaviour or condition that is not genuinely instantiated. Someone may feign illness to deceive others or portray a physician in a theatrical performance. In both cases, simulation involves intentional conduct: the individual knows that an appearance is being produced, whether or not the audience recognizes it as such.

Objective simulation, by contrast, does not require a subject personally to pretend to perform an activity. It involves a system or arrangement capable of reproducing selected properties of a phenomenon or process. The structure of the operation is consequently different. Rather than an individual merely pretending to do something, a device is made to simulate a phenomenon whose behaviour can be observed, controlled, and potentially modified.

This distinction matters because subjective and objective simulation do not have the same relationship to knowledge. Subjective simulation primarily belongs to the domains of pretence, imitation, and performance, whereas objective simulation can serve as a means of investigating processes whose mechanisms, interactions, or possible developments are not directly accessible.

The distinction should not, however, be mistaken for a contrast between subjective and objective experience. It concerns two modes of simulation rather than two degrees of consciousness or authenticity in the phenomena being represented.

Objective simulation provides the appropriate starting point for examining conversational agents. This is not because their status as simulations has already been established, but because the claim that they merely simulate conversation concerns the ability of a computational system to reproduce the properties of a human activity.

An LLM is not an individual who deliberately decides to pretend to converse. It is a technical system that generates responses through computational processes. The relevant question is therefore whether these responses merely reproduce the outward manifestations of conversation or actually perform some of the operations that constitute conversational activity.

The argument from objective simulation rests on an apparently straightforward distinction between reproducing a phenomenon and actually realizing it.

A meteorological model may simulate a storm without producing rainfall. A flight simulator may reproduce the conditions of piloting an aircraft without transporting passengers. In these cases, the simulation reproduces certain properties of a phenomenon without performing the operations that constitute the phenomenon itself.

This distinction between computational simulation and actual realization occupies an important place in philosophical debates concerning artificial intelligence, particularly in John Searle's work \cite{searle1980}.

Yet the distinction becomes problematic when applied to symbolic activities.

A computer performing multiplication does not merely simulate calculation: it actually calculates. Similarly, a computer program participating in a game of chess does not simply reproduce the appearance of playing chess. It makes moves and performs operations within the rules that constitute the game.

Turing's decision to replace the abstract question of whether machines can think with an imitation game organized around observable performance already invites us to reconsider the relationship between internal mechanisms and the accomplishment of an activity \cite{turing1950}. Although Turing's test does not by itself establish the reality of machine understanding, it challenges the assumption that differences in internal constitution necessarily settle questions about what a system can do.

Why, then, should conversation necessarily be different?

To claim that a conversational agent merely simulates human dialogue is to assume that conversation requires subjective properties that the machine lacks. Yet this conclusion presupposes precisely what remains to be demonstrated: that lived intentionality and consciousness are necessary conditions for every genuine conversational activity.

This difficulty lies at the centre of debates concerning artificial understanding, from Searle's Chinese Room argument to contemporary discussions of the relationship between linguistic form, meaning, and communicative intention \cite{searle1980,bender2020}.

Searle's argument challenges the inference from successful symbol manipulation to understanding. A system may produce formally appropriate linguistic responses without possessing the semantic comprehension ordinarily associated with a human speaker. Similarly, Bender and Koller's analysis emphasizes the distinction between learning linguistic form from data and acquiring the communicative meanings to which those forms refer \cite{bender2020}.

These objections establish an important limitation on what may be inferred from linguistic performance. They do not, however, establish that every activity involving the production and interpretation of linguistic signs must be regarded as simulated whenever one participant lacks human-like understanding.

The difficulty arises from conflating the nature of the participants with the nature of their relation.

Conversation is not merely the juxtaposition of two subjectivities. It is constituted through acts of address, response, uptake, and the transformation of utterances within an evolving discursive context.

Herbert Clark and Edward Schaefer's work on discourse contributions and the construction of common ground is particularly relevant here. Their account characterizes conversation as a coordinated activity in which successive contributions modify the shared conditions under which discourse proceeds \cite{clark1989}.

Consider a researcher who submits a hypothesis to an LLM. The system formulates an objection; the researcher challenges that objection; the system subsequently reformulates its argument in response to the criticism. Through this sequence, the conditions of the exchange are effectively modified. A shared discursive object emerges and undergoes transformation.

The relevant phenomenon is not simply the production of an external resemblance to conversation. At least some of the operations that constitute conversational interaction are actually being performed.

This does not imply that the model understands as a human being does or possesses subjective intentionality. But the absence of these properties cannot, by itself, render the conversational operations fictitious.

Chauvier's subsequent reflections on artificial intelligence further underscore the importance of distinguishing the mechanical simulation of intelligent operations from the question of whether particular operations are genuinely accomplished \cite{chauvier2023}.

Objective simulation is therefore a useful category for identifying the reproduction of certain behaviours, but it cannot, on its own, determine whether a symbolic activity is actually taking place.

This becomes especially clear when we consider the difference between producing an isolated conversational-looking utterance and sustaining an interaction in which successive contributions depend on one another.

In the latter case, utterances are not merely generated. They are addressed, received, interpreted, and reintegrated into an evolving relation whose subsequent direction they help determine.

This does not mean that every mediation capable of producing meaning qualifies as conversation. Reading a philosophical book may transform our understanding and prompt objections, but the written text does not change in response to the reader's questions. Interaction with a conversational agent introduces an additional property: a human contribution can condition the generation of a subsequent response, which may in turn alter the direction of the exchange.

The decisive distinction is therefore not simply between a simulated activity and a real one. It is between reproducing selected features of conversational behaviour and actually participating in a developing discursive relation.

The question can no longer be settled solely by examining the internal constitution of the machine. It requires an account of the constitutive properties of conversation and an examination of whether those properties can be realized within a relation between participants of fundamentally different kinds.

This shift opens the way to an understanding of conversational activity grounded not exclusively in the subjective characteristics of interlocutors, but in the technical, relational, and semiotic processes through which discourse is organized.

\section{Conversation as a Relational and Semiotic Reality}

The distinction between objective simulation and actual performance invites a reconsideration of the problem. Rather than asking whether a conversational agent possesses the same properties as a human interlocutor, we must examine what actually occurs during the exchange. Conversation cannot be characterized exclusively through the internal mechanisms of its participants. It is also a temporally organized relational activity in which successive contributions acquire meaning through their connections with preceding and subsequent utterances.

This perspective is consistent with pragmatic approaches to communication, particularly the work of Clark and Schaefer, who understand contributions to discourse as participating in the progressive construction of common ground \cite{clark1989}. Conversation does not consist merely in producing a sequence of grammatically acceptable sentences. Utterances must be addressed to an interlocutor, take account of what has already been said, and remain capable of altering the subsequent course of interaction. Meaning therefore resides not only in the propositions exchanged but also in the ways those propositions are taken up, corrected, challenged, and developed.

From this standpoint, several constitutive properties of conversational activity can be identified: address to an interlocutor, contextual relevance, uptake of previous contributions, the progressive construction of a shared discursive object, and the possibility of redirecting the exchange. These properties do not necessarily exhaust the definition of human conversation. They nevertheless provide criteria for identifying an effective conversational activity that cannot be reduced to the outward imitation of linguistic behaviour.

Consider a researcher who submits to a conversational agent a hypothesis concerning the relationship between intelligence and language. The system raises an objection, the researcher challenges its assumptions, and the agent reformulates its argument in response. After several exchanges, the original problem may have been transformed: a conceptual distinction emerges, an overlooked difficulty becomes explicit, or a new hypothesis is developed.

It would be difficult to maintain that nothing has occurred simply because the conversational agent lacks subjective experience. The discursive interaction has effectively altered the researcher's intellectual activity. The model's responses were not repetitions of a predetermined dialogue; their formulation depended on the particular trajectory of the exchange.

This feature distinguishes conversational interaction from other forms of meaningful mediation. Reading a philosophical work may provoke objections, transform understanding, or generate new insights. Yet the text remains unchanged when the reader questions it. By contrast, in an exchange with an LLM, a human contribution can modify the subsequent response, which then becomes a new condition for the continuation of the discourse. The interaction exhibits a contingency and a capacity for discursive uptake that do not occur in the same way in relation to a fixed text.

Lucy Suchman's work on situated action is particularly instructive in this regard. Rather than deriving human--machine interaction entirely from predefined plans or representations embedded in technical systems, Suchman emphasizes the situated conditions under which activities are actually accomplished \cite{suchman2007}. The point is not to assimilate the capacities of machines to those of human beings. It is to examine how activities take shape within concrete situations involving agents, artifacts, institutional arrangements, and symbolic resources.

A significant objection nevertheless remains. Human conversation is not reducible to the coordination of linguistic behaviours. It also has a normative dimension. When an interlocutor makes an assertion, that person may be required to justify it, acknowledge a contradiction, or answer for the commitments undertaken through speech. In \emph{Making It Explicit} (1994), Robert Brandom provides an influential account of discursive practices structured by the attribution and recognition of commitments and entitlements \cite{brandom1994}.

A conversational agent, however, may produce a justification, acknowledge an error in linguistic form, or revise a proposition without being a subject who bears responsibility for those commitments in the human sense. The difference is fundamental. A response formulated as a promise need not constitute a personal moral commitment. An apology generated by a model does not necessarily express an experience of remorse.

This objection prevents us from simply equating human--machine conversation with interpersonal conversation. Yet it does not establish that the former is fictitious. Rather, it requires a distinction between interactional reciprocity, through which contributions mutually condition one another, and subjective and normative reciprocity, which presupposes capacities and forms of responsibility that cannot automatically be attributed to language models. The actuality of the former does not establish the existence of the latter.

Conversational reality should therefore not be confused with symmetry between interlocutors. A relation can be genuine without its participants possessing identical properties, capacities, or responsibilities. This asymmetry becomes particularly significant when we consider the effects of recognition that conversational agents may elicit.

In ``The Mirror Stage as Formative of the I Function'' (1949), subsequently included in \emph{Écrits} (1966), Jacques Lacan describes the relationship between the subject and its specular image as a decisive mediation in the formation of the ego \cite{lacan1949}. The reflected image possesses no subjectivity of its own. It is not another subject who recognizes the child. Nevertheless, the relationship with that image produces real effects of identification, anticipation, and misrecognition. The reality of these transformations does not depend on the mirror possessing an interior life.

This comparison does not imply that a conversational agent occupies the position of the Lacanian big Other, nor that the processes of identification described by Lacan can simply be transferred to digital interactions. It nevertheless suggests that a mediation devoid of its own consciousness may participate in the ways a subject represents, narrates, and transforms itself.

The concept of the algorithmic mirror, developed in our work on conversational artificial intelligence and the hybrid semiotic fabrication of otherness \cite{denis2026mirror}, designates precisely this configuration. Whereas an optical mirror returns a visual image, a conversational agent generates discursive representations that users may interpret and incorporate into their understanding of themselves. It may reformulate a concern, organize a narrative, propose an interpretation, or attribute coherence to otherwise fragmented statements. The mirror thus acquires a dialogical dimension: rather than merely reflecting, it responds.

Consider an adolescent who confides difficulties concerning relationships to a conversational agent. The system may reformulate the adolescent's experience, identify possible emotional categories, and propose an interpretation of the situation. The effects of such an interaction may be real even though the machine does not experience empathy or concern. Yet this effectiveness may also become problematic. An unfounded interpretation, persistent agreement, or insufficient contradiction may reinforce an inaccurate representation of the self. The issue is therefore not simply whether the agent understands, but how its responses participate in human processes of interpretation and recognition.

This analysis calls for a distinction between subjective otherness, which refers to the lived experience of another subject, and what may be termed functional otherness. A conversational agent need not constitute another subject in order to introduce a genuine difference into an interaction. It may offer an unexpected objection, a reformulation, an unfamiliar perspective, or a discursive resistance to an initial hypothesis. This difference is not imaginary merely because it results from computational processing. It becomes manifest through the transformations it enables within the exchange.

Such an account nevertheless encounters a more radical objection, one that concerns not merely the simulation of behaviour but the possibility that signs might produce a substitute for reality itself. In \emph{Simulacra and Simulation} (1981), Jean Baudrillard develops a conception of the simulacrum that cannot be reduced to the deceptive imitation of an original \cite{baudrillard1981}. For Baudrillard, the simulacrum may designate a regime in which signs and models progressively displace their referents, generating a form of hyperreality whose apparent coherence no longer depends on confrontation with the reality it claims to represent.

The relevance of this perspective to conversational agents is immediately apparent. An artificial system can generate utterances expressing empathy, understanding, uncertainty, or recognition without those linguistic expressions necessarily corresponding to genuinely experienced subjective states. The fluency of the dialogue may encourage users to confuse the appearance of presence with the existence of an interlocutor endowed with an interior life. Machine-generated language may thus sustain a simulacrum of subjective otherness, all the more convincing because it employs the ordinary forms of human communication.

Yet this possibility does not establish that the conversational activity itself is a simulacrum. We must distinguish what the technical system appears to represent from what it actually enables participants to accomplish. A conversational agent may produce the appearance of human subjectivity while participating in a genuine discursive activity. An objection generated by the system may transform a line of reasoning; a reformulation may modify the interpretation of a problem; an unexpected response may lead the user to abandon an initial hypothesis.

These effects do not, by themselves, refute Baudrillard. The simulacrum is not devoid of efficacy. It may generate real social, symbolic, and practical consequences. The decisive distinction must therefore be established elsewhere: in the possibility of differentiating an anthropomorphic representation of the system from the conversational operations that are actually performed.

When a user interprets the agent's expressions of empathy as evidence of a consciousness that experiences them, the representation risks being confused with its supposed referent. When, however, the computational character of the system is acknowledged, the conditions under which its responses are produced remain open to investigation, and its claims can be challenged, the account of conversational reality need not rely on such a confusion. The conversation can be characterized in terms of its effective operations without presupposing the existence of the subjectivity that its linguistic forms may appear to express.

This does not mean that interactions with LLMs are intrinsically immune to the logic of the simulacrum. Certain technical configurations and patterns of use may actively encourage the concealment of mediation and the illusion of human presence. The argument is more precise: the possibility of a simulacrum of subjectivity does not establish that the conversational activity itself is a simulacrum. The actuality of a discursive relation must be distinguished from beliefs concerning the identity or interiority of its participants.

This distinction requires a further philosophical shift. The reality of a relation cannot be identified either with the subjective authenticity of its participants or with the truth of the representations they produce. It must also be understood through the operations, mediations, and transformations that constitute it. Gilbert Simondon's philosophy offers important conceptual resources for developing this relational perspective.

In \emph{L'Individuation à la lumière des notions de forme et d'information}, Simondon challenges approaches that treat individuals as fully constituted entities prior to the relations in which they participate \cite{simondon2005}. Individuation refers to a process through which structures emerge from situations characterized by tensions, potentials, and transformations. It consequently becomes possible to understand an activity not merely as the expression of pre-existing individual properties, but as the result of a relational organization.

It would, however, be unwarranted to conclude that every exchange with an LLM constitutes individuation in Simondon's strict sense, or that a language model undergoes psychic individuation in the same way as a human subject. The relevance of Simondon's philosophy lies elsewhere: it discourages explanations of relations that rely exclusively on the properties of their already constituted terms. In human--machine conversation, the dynamics of the exchange cannot be deduced entirely from the human user considered in isolation, nor from the language model understood as a closed system.

This interpretation is consistent with Simondon's account of technical objects and their conditions of existence \cite{simondon1958}. A technical device is not simply an assembly of components producing predetermined effects. Its functioning must also be understood through the relations it maintains with an associated milieu. Applied to conversational agents, this perspective directs attention not only to the model's parameters but also to the corpora, interfaces, instructions, contexts of use, and human interventions through which conversational activity is concretely organized.

Conversation can thus be approached as a situated reality whose properties arise within an arrangement of heterogeneous mediations. A response is never merely the abstract product of a language model. It occurs within a particular configuration, depends on preceding contributions, and acquires significance through the interpretations and practices to which it gives rise.

Charles Sanders Peirce's semiotics provides a complementary perspective on this process. Within Peirce's triadic account of semiosis, meaning cannot be reduced to a mechanical relationship between a linguistic form and an object. It involves a sign, an object, and an interpretant through which a meaningful relation is constituted and may be further developed \cite{peirce1931}. This account is particularly productive for understanding interactions with LLMs.

The model's outputs function as signs that enter human processes of interpretation. An utterance may elicit an objection, a clarification, a decision, or a new question. Interpretation does not necessarily terminate the process; it may itself become the occasion for further sign production. This dynamic does not require us to attribute human-like subjective understanding to the model. It does, however, require us to recognize that computationally generated utterances and human interpretation can participate in a shared sequence of semiotic activity.

It is precisely this organization that the concept of Hybrid Semiotic Fabrication seeks to describe \cite{denis2025}. Meaning is produced neither by the machine alone nor by a human subject considered independently of technical mediation. It emerges through an arrangement involving human participants, artificial intelligence systems, corpora, and technical devices that condition the production, circulation, and interpretation of signs.

This perspective moves beyond an increasingly inadequate alternative: either artificial systems possess understanding analogous to that of human beings, or their productions are devoid of any meaningful contribution to human intellectual activity. Language models may generate highly coherent utterances without that coherence guaranteeing truth. Their participation in meaning-making can be real without conferring upon them an autonomous epistemic authority.

The distinction between conversational reality and epistemic validity is therefore fundamental. That a conversation genuinely takes place does not imply that its propositions are accurate, its sources reliable, or its interpretations justified. Human conversations, after all, may be entirely genuine while conveying errors, illusions, or fallacious arguments. The existence of a discursive activity and the epistemic standing of its products are distinct questions.

Consider a researcher who asks a conversational agent to identify the sources supporting a scientific claim. The system produces a plausible but nonexistent bibliographic reference. The researcher checks the citation, discovers the error, challenges the response, and obtains a revised answer. The interaction has been genuinely conversational: it has involved a request, a response, an objection, and a revision. Yet its scientific acceptability depends on additional procedures of verification, traceability, and attribution.

This example also extends the discussion of Baudrillard. The risk of the simulacrum emerges when the formal coherence of a response substitutes for verification of its relation to sources and to the world. A fabricated reference may display all the conventional signs of scientific authority without corresponding to any existing document. Hybrid Semiotic Fabrication seeks to preserve the conditions under which this discrepancy can be identified: sources must remain traceable, transformations observable, claims contestable, and errors open to correction.

The aim is not to guarantee that every meaning produced through human--AI interaction escapes the logic of the simulacrum. It is to organize the conditions under which the plausibility of an utterance can be distinguished from its epistemic acceptability. Hybrid Semiotic Fabrication therefore does not entail an uncritical celebration of all outputs generated through human--machine interaction. On the contrary, it requires us to make explicit the conditions under which meanings are elaborated and the criteria by which particular propositions may be accepted, challenged, or rejected.

Such an approach calls for sustained attention to the corpora employed, the provenance of information, the possibility of reconstructing an exchange, and the responsibilities of the human participants involved. We can now understand why describing interactions with artificial agents as merely ``simulated conversations'' is insufficient. That description treats the absence of machine subjectivity as the principal criterion for determining whether conversation takes place. Yet the reality of conversational activity must also be assessed through the relational operations that are actually accomplished.

An artificial agent does not become a human subject by participating in an exchange. But its ontological difference from human beings does not turn the exchange into a fiction. Conversation with a conversational agent is therefore real, although structurally asymmetrical. Its reality lies in the effective organization of discursive operations and in the transformations they make possible, rather than in the presumed existence of artificial consciousness.

This conclusion does not erase what is distinctive about human participation. It makes it possible, instead, to specify the role of human interpretation, judgment, responsibility, and the establishment of conditions under which knowledge claims may be accepted as justified. The conversational agent participates in the elaboration of meanings, but its participation cannot replace the critical requirements through which those meanings acquire epistemic standing.

From this perspective, Hybrid Semiotic Fabrication provides a framework for understanding human--machine conversation as a relational activity whose reality does not depend on the ontological similarity of its participants. It also directs philosophical inquiry toward the technical, semiotic, and normative conditions under which meanings are produced, interpreted, discussed, and validated. Acknowledging the existence of genuine conversation with an artificial agent therefore entails neither attributing human subjectivity to the system nor granting special authority to its outputs. It entails recognizing that discursive operations can be effectively performed within an asymmetrical relation, while preserving the distinction between the processes through which meaning is produced and the conditions under which knowledge claims are justified.

\section*{Conclusion}

The distinction between simulation and conversational reality cannot be settled solely by appealing to the internal properties of artificial agents. It first requires an account of what constitutes conversation and of the operations necessary for its accomplishment. The examination of objective simulation has shown that the computational realization of an activity does not, by itself, establish that the activity is merely simulated. Although certain simulations reproduce the observable properties of a phenomenon without realizing its constitutive operations, this distinction cannot be transferred to symbolic activities without first examining what those activities consist in.

From this standpoint, conversation with an artificial agent can be understood as genuine when it involves effective operations of address, uptake, contestation, and the transformation of meaning. This conclusion does not depend on attributing consciousness or lived intentionality to language models. It follows from recognizing a relational activity whose dynamics cannot be reduced either to the properties of the technical system or to those of the human participant considered in isolation.

The analysis of the algorithmic mirror has illuminated the asymmetrical character of this relation. A mediation devoid of its own subjectivity may nevertheless participate in the ways a human subject represents, interprets, and transforms experience. The confrontation with Baudrillard has also demonstrated that the effectiveness of these transformations does not eliminate the risk of the simulacrum. Conversational agents may encourage misleading representations of understanding, presence, or subjective otherness. It is therefore necessary to distinguish the actuality of conversational operations from the belief that an artificial interlocutor possesses the subjective interiority its language may appear to express.

The perspectives developed by Simondon and Peirce provide further grounds for this distinction by approaching conversation as a relational and semiotic activity. Meaning emerges within situated processes involving human and technical mediations, rather than exclusively from the presumed interiority of interlocutors. This conception does not eliminate the differences between interpersonal and human--machine conversation. It enables those differences to be described more precisely, particularly through the distinction between interactional reciprocity and subjective or normative reciprocity.

The concept of Hybrid Semiotic Fabrication acquires its full significance within this framework \cite{denis2025}. By understanding meaning as the outcome of an arrangement involving humans, artificial intelligence, corpora, and technical devices, it becomes possible to acknowledge the effective contribution of conversational agents to intellectual activity without assigning them autonomous epistemic authority. The reality of an interaction guarantees neither the accuracy of its responses nor the validity of the reasoning it helps produce. It therefore requires renewed attention to the conditions of utterance production, the provenance of information, the practices of interpretation, and the procedures through which claims can be verified and contested.

The philosophical significance of this argument extends beyond the question of whether machines can converse in the same way as human beings. It concerns the broader problem of how intellectual activities should be characterized when they are accomplished within arrangements that associate human participants with technical systems. Recognizing the reality of conversation with an artificial agent requires an account of interaction that does not presuppose ontological similarity between participants. At the same time, this recognition makes it necessary to specify the conditions under which meanings elaborated through such exchanges can become justified knowledge claims.

A relational and semiotic understanding of conversation thus allows us to move beyond the opposition between human authenticity and computational imitation. Rather than asking whether the machine possesses the same subjective properties as a human interlocutor, the central task becomes to determine which conversational operations are effectively realized, how they are technically and socially mediated, and under what conditions their products can be interpreted, evaluated, and justified.


\clearpage
\begin{thebibliography}{99}
\bibitem{turing1950} Turing, A. M. (1950). ``Computing Machinery and Intelligence.'' \emph{Mind}, 59(236), 433--460. \url{https://doi.org/10.1093/mind/LIX.236.433}
\bibitem{chauvier2008} Chauvier, S. (2008). ``Simuler et faire simuler.'' \emph{Revue philosophique de la France et de l'étranger}, 133(3), 279--286.
\bibitem{searle1980} Searle, J. R. (1980). ``Minds, Brains, and Programs.'' \emph{Behavioral and Brain Sciences}, 3(3), 417--424. \url{https://doi.org/10.1017/S0140525X00005756}
\bibitem{bender2020} Bender, E. M., and Koller, A. (2020). ``Climbing towards NLU: On Meaning, Form, and Understanding in the Age of Data.'' In \emph{Proceedings of the 58th Annual Meeting of the Association for Computational Linguistics}, 5185--5198. \url{https://doi.org/10.18653/v1/2020.acl-main.463}
\bibitem{clark1989} Clark, H. H., and Schaefer, E. F. (1989). ``Contributing to Discourse.'' \emph{Cognitive Science}, 13(2), 259--294. \url{https://doi.org/10.1207/s15516709cog1302_7}
\bibitem{chauvier2023} Chauvier, S. (2023). ``IA : le test de la déférence.'' \emph{Revue de métaphysique et de morale}, 119(3), 409--425. \url{https://doi.org/10.3917/rmm.233.0409}
\bibitem{suchman2007} Suchman, L. (2007). \emph{Human-Machine Reconfigurations: Plans and Situated Actions}. Cambridge University Press.
\bibitem{brandom1994} Brandom, R. B. (1994). \emph{Making It Explicit: Reasoning, Representing, and Discursive Commitment}. Harvard University Press.
\bibitem{lacan1949} Lacan, J. (1949). ``Le stade du miroir comme formateur de la fonction du Je telle qu'elle nous est révélée dans l'expérience psychanalytique.'' Reprinted in \emph{Écrits} (1966, pp. 93--100). Seuil.
\bibitem{denis2026mirror} Denis, C. (2026). ``Du stade du miroir au miroir algorithmique : intelligence artificielle conversationnelle et fabrique sémiotique hybride de l'altérité'' [From the Mirror Stage to the Algorithmic Mirror: Conversational Artificial Intelligence and the Hybrid Semiotic Fabrication of Otherness]. \emph{Nouvelle Revue de l'Enfance et de l'Adolescence}, forthcoming. \url{https://hal.science/hal-05640432v2}
\bibitem{baudrillard1981} Baudrillard, J. (1981). \emph{Simulacres et simulation}. Galilée. Especially ``La précession des simulacres.''
\bibitem{simondon2005} Simondon, G. (2005). \emph{L'Individuation à la lumière des notions de forme et d'information}. Jérôme Millon.
\bibitem{simondon1958} Simondon, G. (1958). \emph{Du mode d'existence des objets techniques}. Aubier.
\bibitem{peirce1931} Peirce, C. S. (1931--1958). \emph{Collected Papers of Charles Sanders Peirce} (8 vols.; C. Hartshorne and P. Weiss, Eds., Vols. 1--6; A. W. Burks, Ed., Vols. 7--8). Harvard University Press. See especially CP 2.228.
\bibitem{denis2025} Denis, C. (2025). ``Vers une coconstruction harmonieuse du sens entre langage humain et machinique au sein de la fabrique sémiotique hybride'' [Towards the Harmonious Co-construction of Meaning between Human and Machine Language within Hybrid Semiotic Fabrication]. \emph{Socio-anthropologie}, 52, 37--50. \url{https://doi.org/10.4000/15qda}
\end{thebibliography}
\end{document}